\pdfoutput=1

\documentclass[11pt]{article}

\usepackage[preprint]{acl}

\usepackage{times}
\usepackage{latexsym}

\usepackage[T1]{fontenc}

\usepackage[utf8]{inputenc}

\usepackage{microtype}

\usepackage{inconsolata}

\usepackage{graphicx}
\usepackage{arydshln}
\usepackage{booktabs}
\usepackage{multicol}
\usepackage{amsmath}
\usepackage{amssymb}
\usepackage{multirow} 
\usepackage{tikz} 
\usepackage{pgfplots}
\usepackage{color}
\pgfplotsset{compat=1.18}

\usetikzlibrary{patterns}

\newcommand*{\tien}{\textcolor{black}}

\newcommand*{\phuong}{\textcolor{black}}
\definecolor{gray1}{rgb}{0.95, 0.95, 0.96}
\definecolor{darkgreen}{RGB}{1,50,32}
\definecolor{ForestGreen}{RGB}{34,139,34}
\title{Non-Interactive Symbolic-Aided Chain-of-Thought\\ for Logical Reasoning}

\author{
 \textbf{Phuong Minh Nguyen}\and
 \textbf{Tien Huu Dang}\and
 \textbf{Naoya Inoue}
\\
\\
  Japan Advanced Institute of Science and Technology
\\
\texttt{\{phuongnm,tiendh,naoya-i\}@jaist.ac.jp}
\\
 \small{
   \textbf{Correspondence:} \href{mailto:phuongnm@jaist.ac.jp}{phuongnm@jaist.ac.jp}
 }
}

\begin{document}
\maketitle
\begin{abstract}

\tien{This work introduces Symbolic-Aided Chain-of-Thought (CoT), an improved approach to standard CoT, for logical reasoning in large language models (LLMs).}
The key idea is to integrate lightweight symbolic representations into few-shot prompts, structuring the inference steps with a consistent strategy to make reasoning patterns more explicit within a \phuong{non-interactive} reasoning process.
By incorporating these symbolic structures, Symbolic-Aided CoT preserves the generalizability of standard prompting techniques while enhancing the transparency, interpretability, and analyzability of LLM logical reasoning. 
Extensive experiments on four well-known logical reasoning benchmarks---ProofWriter, FOLIO, ProntoQA, and LogicalDeduction, \tien{which cover diverse reasoning tasks and scenarios}---demonstrate the effectiveness of the proposed approach, particularly \tien{in} complex reasoning tasks that require navigating multiple constraints or rules. 
Notably, Symbolic-Aided CoT consistently improves \tien{LLMs' reasoning capabilities across various model sizes} 
and significantly outperforms conventional CoT on three out of four datasets, ProofWriter, ProntoQA, and LogicalDeduction.
\end{abstract}
 
\section{Introduction} 
In recent years, pre-trained Large Language Models (LLMs) have achieved exceptional success across a wide spectrum of Natural Language Processing (NLP) tasks~\cite{NEURIPS2022_9d560961, shin-van-durme-2022-shot, llama3, yang2025qwen3}. \tien{As a result, LLMs have become a central paradigm} 
in NLP research and applications. 
Their impressive performance is largely attributed to their ability to perform few-shot in-context learning—the mechanism by which models infer solutions based solely on the format and structure of the input prompt, without requiring gradient computations~\cite{NEURIPS2020_1457c0d6,garg2022can,NEURIPS2022_9d560961}.  
\begin{figure}[!htbp]
    \centering 
    \includegraphics[width=\linewidth, keepaspectratio, 
            trim={0 0 0 0 }, page=1, clip=true]{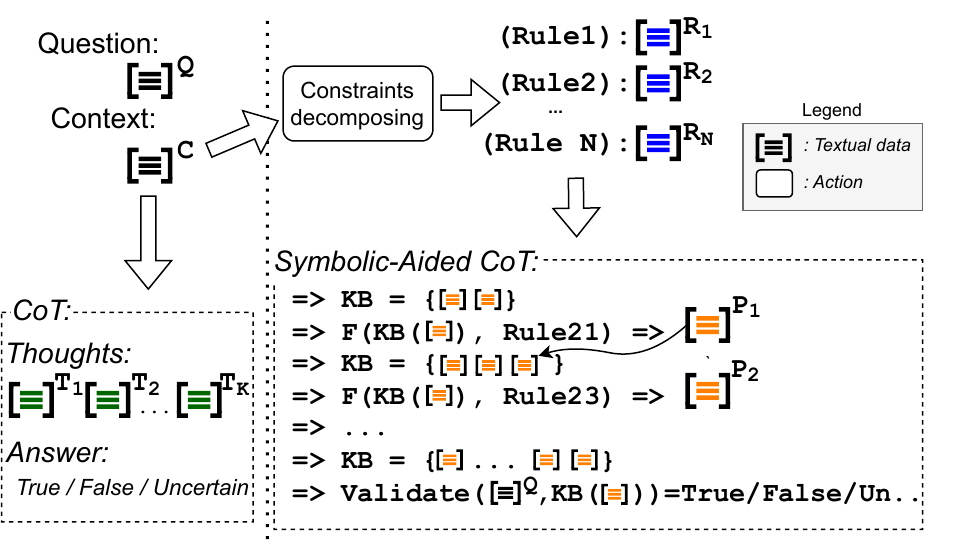}
   
    \caption{Comparison \tien{between standard} CoT and Symbolic-Aided CoT \tien{for logical reasoning tasks}. 
    }\label{fig_idea}
\end{figure}

Notably, as model size grows, prompting methods that leverage intermediate reasoning steps consistently surpass standard input–output prompting methods. 
This reasoning strategy, known as Chain-of-Thought prompting (CoT;~\citet{NEURIPS2022_9d560961}), relies on explicitly modeling the reasoning process. 
For in-context learning, the CoT reasoning technique has demonstrated compelling results across a variety of NLP tasks \cite{NEURIPS2022_9d560961,zhou2023leasttomost}.
Despite recent advancements, applying LLMs to logical reasoning tasks still faces several critical challenges, including conflicts between pretrained knowledge and counterfactual assumptions~\citep{ortu-etal-2024-competition}, failures on cyclic inference graphs~\citep{zhang2025cumulative}, and planning errors during the solving process~\cite{ye2023satlm}. 

To address these issues, \tien{various strategies have been proposed, which can be broadly categorized into two main} groups:
(1) designing an external symbolic solver, which delegates the actual reasoning process to an automated theorem prover~\cite{ye2023satlm,pan-etal-2023-logic} or programming languages~\cite{pmlr-v202-gao23f,xu-etal-2024-symbol}; and (2) constructing a framework that systematically decomposes complex tasks into subtasks---such as rule selection, premise inference, and scoring—to enhance the overall reliability of the system~\citep{zhang2025cumulative,feng-etal-2024-language,sun-etal-2024-determlr,xu-etal-2024-faithful}.
Although the first approaches potentially achieve remarkable performance, they typically require powerful LLMs~\cite{ye2023satlm,pmlr-v202-gao23f} such as GPT-4~\cite{achiam2023gpt} or additional extensive pre-training phase~\cite{xu-etal-2024-symbol,feng-etal-2024-language} for successful parsing from problem description to the logical forms (\textit{e.g.,} First-order logics - FOL). 

In this study, we focus on the second approach, which aims to improve LLM logical reasoning \textbf{\textit{without relying on any external symbolic reasoner or programming language}}. Building on insights from recent works~\cite{sun-etal-2024-determlr,qi2025large,zhang2025cumulative}, we target the challenge of designing a mechanism to systematically decompose complex logical reasoning tasks into simpler subtasks that can be solved by the inherent understanding ability of LLMs in a single inference pass. Closely related to our work, \citet{xu-etal-2024-symbol,feng-etal-2024-language} recently introduced LLM-based frameworks that leverage first-order logic---\textit{a strict formal language} with well-defined syntax and semantics---to support faithful logical reasoning. However, previous studies~\cite{sun-etal-2024-determlr,xu-etal-2024-faithful,zhang2025cumulative} primarily rely on self-refinement or interactive (multi-turn) generation, where each turn solves a sub-task and its output is passed to the next. They overlook non-interactive reasoning, in which the LLM performs the entire reasoning process without any assistance from external modules or sub-processes. To address this gap, we explore \textit{\phuong{non-interactive (single-turn) reasoning generation}}, allowing a more direct and fair comparison with CoT prompting.

We introduce Symbolic-Aided CoT, a novel variant of CoT~\cite{NEURIPS2022_9d560961}, designed to leverage symbolic representations to enhance the \phuong{non-interactive} logical inference capabilities of LLMs (Figure~\ref{fig_idea}). \phuong{In conventional CoT, intermediate reasoning steps are provided in few-shot demonstrations as unstructured text, enabling LLMs to approximate the logical reasoning process. However, relying solely on textual descriptions for complex reasoning introduces ambiguity, as the inherent vagueness of natural language limits LLMs’ ability to generalize precise reasoning steps.} Our core idea is to integrate \textit{lightweight} symbolic structures into few-shot prompts, making the inference steps more transparent and structured, while simultaneously strengthening the induction-head behavior of LLMs~\cite{olsson2022context}.
Specifically, our Symbolic-Aided CoT prompting framework explicitly outlines essential reasoning substeps: \textit{rule matching}---selecting constraint rules that align with the current state of inference, \textit{new premise inference}---applying the selected rules to generate new premises, and \textit{knowledge base (KB) updating}---appending the newly inferred premises to the KB.
By \tien{incorporating} these symbolic structures, our method preserves the flexibility and general applicability of standard prompting techniques while enhancing both the interpretability and analyzability of LLM reasoning behavior. Empirical evaluations across four reasoning QA benchmarks, ProofWriter, FOLIO, ProntoQA, and LogicalDeduction, demonstrate the effectiveness of our approach, particularly in scenarios involving complex reasoning that requires navigating multiple rules and constraints. Remarkably, our Symbolic-Aided CoT, when applied with open-source LLMs (\textit{e.g.}, Qwen3), achieves performance comparable to that of a complex multi-turn reasoning framework~\cite{sun-etal-2024-determlr} built on the powerful GPT-4 model, particularly on the ProofWriter, ProntoQA, and LogicalDeduction datasets.

The remainder of this paper is organized into five sections. Section~\ref{sec_preliminary} provides an overview of logical reasoning tasks and compares our approach with prior studies. Section~\ref{sec_method} presents the details of our proposed framework and its variants. Section~\ref{sec_result} describes the experimental setup and reports the results, with key findings discussed in Section~\ref{sec_discusion}. Finally, the conclusions summarize our contributions and outline directions for future work are presented in Section~\ref{sec_conclusion}.
\section{Background and Related Work\label{sec_preliminary}} 
\subsection{Background}
\paragraph{Notation and problem formulation.} Logical reasoning is a fundamental NLP task within the Question Answering domain \cite{weston2015towards,tafjord-etal-2021-proofwriter}. In this task, machine learning models are required to answer a question based on a context containing multiple logical conditions or constraints. Formally, we denote the list of rules (or constraints) provided in the context as $\mathcal{R} = \{r_i\}_{0 \leq i < N}$. Given a question $(Q)$, the correct answer $(A)$ must be derived from knowledge supported by a subset of rules $\mathcal{R}^* \subset \mathcal{R}$.
To address the challenges posed by logical reasoning, prior research largely falls into two overarching directions: utilizing \textit{external symbolic solvers} and developing \textit{LLM-based logical solvers}.

In the first approach, the main idea is to leverage LLMs to translate textual descriptions of constraints and the question into formal logical formulas, which are then passed to an explicit symbolic solver (\textit{e.g.,} the Z3 theorem prover\footnote{\url{https://github.com/Z3Prover/z3}}) to derive the final answer. Formally, all constraints are aggregated to construct a logical program: $\mathcal{F} = \{  \mathrm{LLM}^{\textrm{lang2logic}}(r_i)\} _{ r_i \in \mathcal{R}}$. Similarly, the question is also transformed to the logical form $f^q = \mathrm{LLM}^{\textrm{lang2logic}}(Q)$. Then, a symbolic reasoner is reasoned over the transformed logical forms to yield the final answer: $\mathrm{SymbolicReasoner}(\mathcal{F}, f^q)$. 

In the second approach, the original logical reasoning problem is decomposed into a sequence of subtasks. Each subtask is solved individually, and the process iterates over multiple turns until a specified stopping condition is satisfied, ultimately producing the final result: $\mathrm{Loop ([\mathrm{LLM}^{\textrm{subtask}_t}(\mathcal{R}, Q)]_t)}$ where $\textrm{subtask}_t$ may represent an arbitrary LLM-based unit function such as \textit{rule matching}, \textit{rule inference}, or \textit{new premise scoring}, among others. These sub‑tasks are typically carefully designed and arranged with sequential dependencies, with the aim of mitigating hallucinations.

\subsection{Related Works\label{sec_relatedworks}}

Building on the success of LLMs across a wide range of NLP tasks~\cite{JMLR:v25:23-0870,llama3,yang2025qwen3}, logical reasoning has emerged as a particularly important area of study, serving as a key benchmark for evaluating both the reasoning capabilities and the overall intelligence of these models~\cite{zhou2023leasttomost,feng-etal-2024-language}. As aforementioned, we categorize recent work into two main approaches:

\paragraph{Utilizing external symbolic solvers.}  In this line of work, the main challenge lies in transforming the description of constraints into logical formulas, $\mathrm{LLM}^{\textrm{lang2logic}}$, effectively functioning as a semantic parser. More specifically, \citet{yao2023react} introduced a method to enrich CoT prompting by leveraging results from external APIs, invoking functions to retrieve supplementary information. Building on this idea, \citet{pmlr-v202-gao23f} proposed an approach that augments the intermediate reasoning steps of CoT reasoning through a runtime environment (\textit{e.g.,} a Python interpreter), which has proven particularly effective for mathematical reasoning tasks. Further, \citet{pan-etal-2023-logic,ye2023satlm,olausson2023linc} enhanced the performance of logical reasoning by translating all constraints in a sample into logical forms, completing a logical program, and solving it using an independent symbolic reasoner. In addition, \citet{xu-etal-2024-symbol} strengthened the logical‑form parsing process of the open-source LLMs by leveraging fine-tuning on a large‑scale, curated dataset. In contrast, our method does not rely on any external symbolic solver; instead, it integrates symbolic syntax directly into the reasoning process, aiming to enable the LLM itself to reason as a symbolic solver.

\paragraph{LLM-based logical solvers.} This approach leverages LLMs directly by designing frameworks that decompose complex reasoning tasks into smaller, manageable subtasks—such as rule selection, premise derivation, and scoring—to enhance overall reasoning robustness~\citep{zhang2025cumulative,feng-etal-2024-language,sun-etal-2024-determlr}. In particular, \citet{sun-etal-2024-determlr,xu-etal-2024-symbol} introduced a framework that enables LLMs to uncover hidden premises and integrates scoring components through a multi‑turn reasoning process. Similarly, \citet{zhang2025cumulative} proposed a framework that incrementally refines reasoning via three subcomponents: Proposer, Verifier, and Reporter. Finally, \citet{feng-etal-2024-language} presented LoGiPT, a method that enhances LLMs’ ability to function as logical solvers by learning the reasoning process step by step through additional training on large‑scale data collected from the reasoning traces of external symbolic solvers. Compared to our work, LoGiPT similarly performs step‑by‑step reasoning as a logical solver and can produce a proof tree at the end; however, \textit{our method achieves competitive performance without the need for any additional training.}


\section{Methodology\label{sec_method}}
In this section, we present our \textit{Symbolic‑Aided CoT} method, its variants, and the motivation behind it in comparison with baseline prompting techniques: \textit{Standard} - which directly provides an answer without any reasoning - and \textit{CoT} \citep{NEURIPS2022_9d560961} - which produces an answer accompanied by step‑by‑step reasoning. All of these prompting methods are augmented with a hard-selected few‑shot examples included in the prompt \cite{qi2025large}.
All prompts are directly fed forward through the LLM to obtain the final predicted answer in a single turn:
    \begin{align}
        A^{\text{out}} =  \mathrm{LLMs}( \mathrm{prompting} (\mathcal{R}, Q))
    \end{align}
In this setup, the LLM is solely responsible for generating the desired answer given the contextual input. The \textrm{prompting} component consists of a few‑shot template designed to help the LLM understand the task description while leveraging its own knowledge to reason over the list of provided rules in the contextual information. For clarity, the templates for \textit{Standard} and \textit{CoT} prompting are shown in the first two rows of  Table~\ref{tab_prompting}.
        \begin{table}[!htbp]
            \small
            \centering
            \caption{Template of input and output for prompting techniques: \textit{Standard} and \textit{CoT} and \textit{Symbolic-Aided CoT}.\vspace{-15pt}
            \label{tab_prompting} }
            \resizebox{\linewidth}{!}{%
                \begin{tabular}{|p{0.14\linewidth}p{0.9\linewidth}|lrcrr}
                    \hline
                Standard  \newline\textit{(Input)} 
                    & Context: \textit{[[All constraints, $\mathcal{R}$]]} \newline
                     Question: \textit{[[Content of the question, $Q$]]} \newline
                     Options: A) True B) False C) Uncertain 
                    \\[3pt]
                Standard\newline\textit{(Output)} 
                    &  The correct option is:   \newline\{  "answer": \textit{[[$A$]]}  \}\\\hline\hline
                CoT \newline\textit{(Input)} 
                    & Context: \textit{[[All constraints, $\mathcal{R}$]]} \newline
                     Question: \textit{[[Content of the question, $Q$]]} \newline
                     Options: A) True B) False C) Uncertain 
                    \\[3pt]
                CoT  \newline\textit{(Output)} 
                    &  The correct option is: \newline\{\newline"reasoning": \textit{[[reasoning content]]} , \newline "answer": \textit{[[$A$]]} \newline\} \\\hline\hline
                Symbolic-Aided \newline CoT  \newline\textit{(Input)} 
                    & \#\#\# Let us define F as a function that infers new premises based on a given list of facts and rules. Using these facts and rules, provide a reasoning path that leads to one of the values of a Validate function: True, False, or Uncertain.  \newline
                    ------\newline
                    \#\#\# Example1: Given list of facts and rules:\newline 
                    \noindent\colorbox{gray1}{\fbox{\parbox{0.97\linewidth}{%
                        \# (Rule[[$i$]]): \textit{[[content of $r_i \in \mathcal{R}$]]} $\cdots$ 
                    }}}
                    \newline 
                    \# (Question): \textit{[[content of the question, $Q$]]}
                    \\[3pt]
             Symbolic-Aided \newline CoT  \newline\textit{(Output)} 
                    &  \# (Answer): Start from the object and their condition mentioned in the question to collect relevant facts:  \newline
                        \# KB = \{ \} \newline
                    \noindent\colorbox{gray1}{\fbox{\parbox{0.97\linewidth}{%
                        => F(KB(\textit{[[premises in KB]]}), Rule$[[i']]$) => \textit{[[inferred premises]]} \newline
                        \# KB = \{\textit{[[KB values for each reasoning steps]]}\} 
                        }}}
                    \newline 
                    \# validate the question with the current inferred premise \newline
                    => Validate(Question=\textit{[[$Q$]]}, KB(\textit{[[selected premise]]})) = \textit{[[$A$]]}.
                    \\  
                      \hline
                \end{tabular} 
            } 
        \end{table}
\subsection{Symbolic-Aided CoT }
We formulate logical reasoning tasks into three fundamental sub‑tasks, namely, reasoning operators: \textit{rule matching}, \textit{rule inference}, and \textit{knowledge base updating}. Previous frameworks, such as DetermLR \cite{sun-etal-2024-determlr} and CR \cite{zhang2025cumulative}, were also built on carefully designed unit operators, integrating them with procedural programming to process the outputs of LLMs. The key difference between our Symbolic‑Aided CoT and these approaches is that Symbolic‑Aided CoT is conceived entirely as an LLM‑driven program. In our design, the LLM is expected to learn the flow of the logical reasoning process from a few‑shot demonstration. To this end, the LLM has full visibility of all sub‑reasoning steps and autonomously decides which step to execute next. The overview of our Symbolic‑Aided CoT is presented in the third row of Table~\ref{tab_prompting}, which illustrates the instruction text, list of rules, question, and reasoning‑flow examples. For a clearer explanation, we elaborate on the two gray blocks shown in this table, which pertain to rule tagging and reasoning operators in our Symbolic‑Aided CoT. 
        
\paragraph{Rule tagging.} In preparing the prompt input, we first segment the contextual information into a list of rules by splitting it into individual sentences using the NLTK toolkit\footnote{\url{https://www.nltk.org/}}. Each sentence is then tagged with its order index (\textit{e.g.,} \texttt{Rule5} for the fifth sentence), allowing the LLM to track and reference the reasoning steps. Here, we assume that the LLMs can link each rule’s content to its corresponding tag and reference this symbol appropriately in the reasoning flow in subsequent steps.

\begin{figure}[!htbp]
    \centering 
    \includegraphics[width=\linewidth, keepaspectratio, 
            trim={0 0 0.5cm  0}, page=1, clip=true]{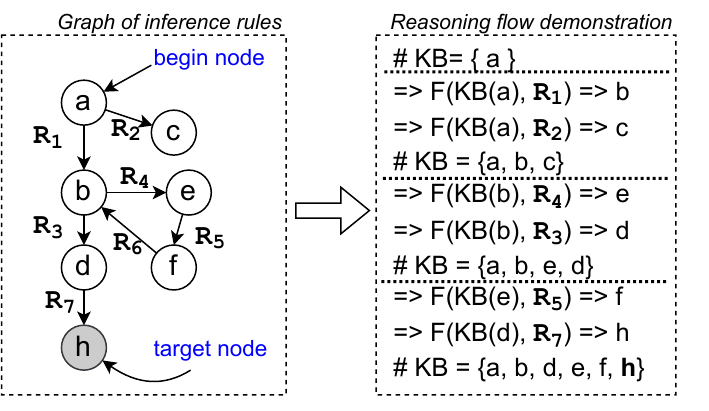}
    \caption{\textbf{Left}: Graphical model of inference rules. \textbf{Right}: Reasoning flows in the Symbolic‑Aided CoT demonstrations.}\label{fig_reasoningflow}
\end{figure}

\paragraph{Reasoning operators.} This demonstration serves as the core example that enables LLMs to learn, in context, the pattern for solving logical reasoning tasks. We use a set of symbols, similar to those in programming languages, to represent the inference flow (see Figure~\ref{fig_reasoningflow}). At each reasoning step, the LLM selects the relevant rules and premises from the current knowledge base (KB) to infer new premises. Each newly inferred premise is then appended to the KB for use in the next inference step. A breadth‑first search strategy is applied to traverse the nodes (premises), as illustrated in Figure~\ref{fig_reasoningflow}. Inspired by how humans solve such tasks, we maintain a KB state to prevent cyclical inference loops: if a newly inferred premise already exists in the KB, it is not added again. All patterns of rule selection, inference, and KB updating are implicitly conveyed within the demonstrations provided in the few‑shot prompts, allowing the LLM to internalize these reasoning steps.   

\section{Experiment and Analysis\label{sec_result}}
    In this section, we present a detailed description of our experiments, together with the results and analyses, to assess the effectiveness of our Symbolic-Aided CoT prompting in comparison to standard CoT and prior methods.
    \subsection{Experimental Setup}
        \paragraph{Datasets.} We conducted experiments on four well-known benchmark datasets about the logical reasoning task: (1) \textbf{ProofWriter}~\cite{tafjord-etal-2021-proofwriter} - we use the subset under the open-world assumption, where each sample has three possible answer options: true, false, or unknown. Following~\citet{pan-etal-2023-logic}, we evaluate on the subset with the longest reasoning depth (5 hops), which contains $600$ cases. (2) \textbf{FOLIO}~\cite{han-etal-2024-folio} is a challenging, expert-curated dataset for logical reasoning that contains rules closely aligned with real-world knowledge. Following the setup of previous work~\cite{sun-etal-2024-determlr}, we evaluate our method on a subset of this dataset comprising $204$ examples.  (3) \textbf{ProntoQA} introduced by \citet{saparov2023language} - similar to the ProofWriter dataset, we also choose the hardest subset of this data with $5$-hop reasoning across $500$ samples for the evaluation, following previous works~\cite{sun-etal-2024-determlr,qi2025large}; (4) \textbf{LogicalDeduction}~\cite{srivastava2023beyond} is a dataset for logically identifying the order of objects given a list of description constraints. We follow the previous setting from~\citet{sun-etal-2024-determlr}, using $300$ evaluation samples containing all subsets of three, five, and seven objects (the greater the number of objects, the more complex the logical reasoning required to determine their order).
        
        \paragraph{Evaluation metric.} In order to evaluate system performance, we use Accuracy as the metric, which is standard and allows for direct \tien{and fair} comparison with previous works~\citep{pan-etal-2023-logic,sun-etal-2024-determlr,qi2025large}.

        \paragraph{Setting.} We conducted our experiments primarily on open-source LLMs, including \texttt{Llama-3.1-8B-Instruct}~\cite{grattafiori2024llama} and the \texttt{Qwen-3} models~\cite{yang2025qwen3}. Specifically, we aim to evaluate the effectiveness of our Symbolic-Aided CoT compared to standard CoT and standard prompting on these LLMs. In addition, we performed extensive experiments on the \texttt{Qwen-3} models across various sizes---\texttt{1.7B}, \texttt{4B}, \texttt{8B}, and \texttt{32B}---to further assess the scalability and effectiveness of our proposed prompting method. Finally, for comparison with previous approaches, we also conducted extensive experiments on a powerful closed-source LLM, \texttt{GPT-4}~\cite{achiam2023gpt}, to evaluate our proposed method. For open-source models, we use greedy decoding to generate the answer \textit{i.e.,} the token with the maximum logit is picked.

    \subsection{Main Results}
       For the main results across all benchmark datasets, ProofWriter, FOLIO, and ProntoQA, we present the performance of our Symbolic-Aided CoT compared with Standard Prompting, CoT Prompting, and previous methods in Table~\ref{tab_mainresult}. 
       
       \paragraph{Overall performance.} These results demonstrate the clear superiority of our proposed method over the standard CoT on three datasets, ProofWriter, ProntoQA, and LogicalDeduction. Notably, on the ProofWriter dataset, Symbolic-Aided CoT significantly outperforms CoT, achieving improvements of $15\%$ and $21\%$, $22\%$ on the \texttt{Qwen-3-8B}, \texttt{Qwen-3-14B}, and \texttt{Llama3.1-8B-Instruct} models, respectively. In addition, the improvement is clearly observed on the LogicalDeduction and ProntoQA datasets across three open-source LLMs.
        These findings further highlight that the degree of improvement varies across different LLMs. The effectiveness of Symbolic-Aided CoT largely depends on each model’s ability to understand logical relationships and recognize logical matching patterns embedded within the few-shot prompts. Moreover, our method is simple yet effective, achieving competitive performance compared to prior works such as Logic-LM~\cite{pan-etal-2023-logic} and DetermLR~\cite{sun-etal-2024-determlr}, even when those methods are supported by a more powerful \texttt{GPT-4} model. 
        
        On the FOLIO dataset, the results show that the Symbolic-Aided approach has a weakness compared to CoT prompting, especially with the \texttt{Qwen-14B} LLM.
        We found that the FOLIO dataset is specifically designed by experts to cover various aspects of factual knowledge, which allows the CoT prompting technique to leverage this advantage (leaking factual knowledge) when solving questions.
        For example, in a question about the tennis player Djokovic, CoT prompting tends to use external knowledge such as \textit{``Djokovic is famous and is an athlete'' }, which is not provided in the set of facts in the context, to support the inference flow. In contrast, our Symbolic-Aided CoT approach relies strictly on the inference rules given in the context. 

        Finally, we evaluate our proposed method in the setting that uses \texttt{GPT-4} as the backbone LLM for the reasoning task (last row of Table~\ref{tab_mainresult}). \phuong{Compared to the SymbCoT framework~\cite{xu-etal-2024-symbol}, our method achieves superior performance on ProntoQA but lower performance on other datasets. This difference can be attributed to SymbCoT’s use of complex interactive reasoning sub-steps---such as translator, planner, solver, and verifier---each supported by carefully designed prompts tailored to the specific sub-step and logical reasoning task.}  Overall, our method surpasses the performance of previous methods on the ProntoQA dataset and achieves remarkable results on the ProofWriter, FOLIO, and LogicalDeduction datasets. These results demonstrate the robustness of our approach, even in the stringent setting that uses only \phuong{non-interactive} inference without the support of an external solver or multi-step inferences, such as the DetermLR~\cite{sun-etal-2024-determlr} or LogicLM~\cite{pan-etal-2023-logic} approaches. 

     \begin{table*}[!htbp]
            \small
                \caption{{Performance comparison among different methods. \textit{E.Solver} refers to the system supported by an external symbolic solver module. \textit{ICL} stands for in-context learning, and \textit{Supervised FT} stands for the supervised fine-tuning approach. \tien{The \textbf{best} is marked.} }\label{tab_mainresult}}
            \centering
            \resizebox{\textwidth}{!}{%
                \begin{tabular}{lllrrrrr}
                    \toprule  \multirow{1}{*}{\textbf{Methods}}&\textbf{Learning paradigm}&\textbf{Interaction mode }&\textbf{ProofWriter}&\textbf{FOLIO} &\textbf{ProntoQA} &\textbf{L.Deduction}  \\
                    \midrule
                     \multicolumn{4}{l}{\quad \quad\underline{\textit{Llama3.1-8B-Instruct}}} \\
                     
                    Fine-tuned ID   \cite{qi2025large} & Supervised FT & Non-Interactive  &  $71.67$ & $70.00$ &  --- & ---\\\hdashline
                    Standard  & ICL few-shot& Non-Interactive &  $36.83$ & $53.92$ & $51.80$ & $40.67$\\
                    CoT  &ICL few-shot &Non-Interactive &  $44.83$&   $\mathbf{56.86}$& $74.00$ & $58.00$	\\  
                    \textbf{Symbolic-Aided CoT (ours)} & ICL few-shot &Non-Interactive & $\mathbf{68.67}$ &  $55.88$& $\mathbf{89.00}$ & $\mathbf{59.33}$\\ 
                     \midrule
                     \multicolumn{4}{l}{\quad \quad\underline{\textit{Qwen3-8B}}} \\
                    Standard  & ICL few-shot&Non-Interactive & $60.00$ & $62.25$ & $80.80$ & $57.00$\\ 
                    CoT & ICL few-shot& Non-Interactive& $57.83$& $\mathbf{66.67}$ & $95.80$	& $72.67$ \\ 
                    \textbf{Symbolic-Aided CoT (ours)} & ICL few-shot&Non-Interactive & $\mathbf{78.67}$ & $65.69$ & $\mathbf{97.20}$& $\mathbf{77.33}$\\   
                     \midrule
                     \multicolumn{4}{l}{\quad \quad\underline{\textit{Qwen3-14B}}} \\
                    Standard  &ICL  few-shot&Non-Interactive & $46.50$ & $67.16$ &$ 77.80$&$ 61.67$\\
                    CoT  &ICL few-shot&Non-Interactive &  $62.67$& $\mathbf{74.02}$ & $97.20$ & $81.67$\\ 
                    \textbf{Symbolic-Aided CoT (ours)} &ICL few-shot&Non-Interactive &  $\mathbf{77.00}$ &  $65.20$  & $\mathbf{97.80}$& $\mathbf{86.33}$\\
                    \midrule
                     \multicolumn{4}{l}{\quad \quad\underline{\textit{GPT-4}}} \\
                     
                    CoT~\cite{sun-etal-2024-determlr}  &ICL few-shot& Non-Interactive & $67.41$ &  $67.65$ & $91.00$ & $73.33$ \\
                    Logic-LM~\cite{pan-etal-2023-logic}  &ICL few-shot & Interactive +E.Solver & \underline{$79.66$} & \underline{$78.92$} & $83.20$& \underline{$87.63$}\\
                    DetermLR~\cite{sun-etal-2024-determlr}  &ICL few-shot& Interactive +Programming& $79.17$ & $75.49$ & $98.60$ & $85.00$\\
                    SymbCoT~\cite{xu-etal-2024-symbol}  &ICL few-shot & Interactive +Programming & $\mathbf{82.50}$ & $\boldsymbol{83.33}$ & \underline{$99.60$}&$\mathbf{93.00}$ \\
                    \textbf{Symbolic-Aided CoT (ours)} & ICL few-shot &Non-Interactive & $77.09$  &  $74.51$ & $\boldsymbol{100.00}$ & $86.33$\\ 
                    \bottomrule 
                \end{tabular} 
                } 
        \end{table*}
        
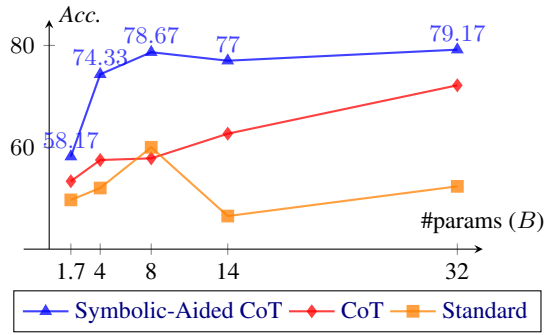
\begin{figure}[ht]
            \centering
            \footnotesize
            \begin{tikzpicture}
                \begin{axis}[width=0.99\linewidth,height=0.6\linewidth,
                    axis lines=middle,
                    ymin=40, ymax=85,
                    xmin=-2, xmax=34, 	          
                    x label style={at={(current axis.right of origin)},anchor=south, above=1mm},
                    legend style ={at={(0.7,0.8)}, anchor=north, align=left},
                    legend cell align={left},
                    legend style={ anchor=north,legend columns=-1},
                    legend to name={legend_ratio},
                    xlabel= \#params ($B$),
                    ylabel=\textit{Acc.},
                    ylabel style = {at={(0.06,1.1)},},
                    xticklabel,  
                    enlargelimits = false,
                    xticklabels from table={./model_size_ab.tex}{depth},
                    every axis plot/.append style={ thick},
                    xtick=data]

                    \addplot [ opacity=0.75,mark=triangle*,
                   nodes near coords, color=blue,  error bars/.cd, y dir=both, y explicit,] table [ x=idx, y=LogicCoT, col sep=space, 
                        ]   {./model_size_ab.tex};
                    \addlegendentry{Symbolic-Aided CoT} 
                    \addplot [ opacity=0.75,mark=diamond*, color=red,  error bars/.cd, y dir=both, y explicit,] table [ x=idx, y=CoT, col sep=space, 
                        ]   {./model_size_ab.tex};
                    \addlegendentry{CoT} 
                    \addplot [ opacity=0.75,mark=square*, color=orange,  error bars/.cd, y dir=both, y explicit,] table [ x=idx, y=Standard, col sep=space, 
                        ]   {./model_size_ab.tex};
                    \addlegendentry{Standard} 
                    
                \end{axis} 
            \end{tikzpicture}
            \ref{legend_ratio}
            \caption{Performance across different model sizes of \texttt{Qwen-3} with three prompting techniques on the ProofWriter dataset. 
            \label{fig_modelsize}}
        \end{figure}

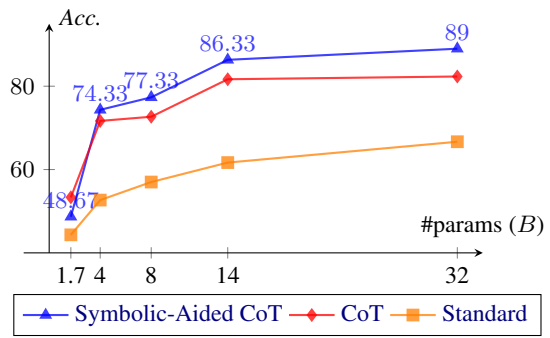
\begin{figure}[ht]
            \centering
            \footnotesize
            \begin{tikzpicture}
                \begin{axis}[width=0.99\linewidth,height=0.6\linewidth,
                    axis lines=middle,
                    ymin=40, ymax=95,
                    xmin=-2, xmax=34, 	          
                    x label style={at={(current axis.right of origin)},anchor=south, above=1mm},
                    legend style ={at={(0.7,0.8)}, anchor=north, align=left},
                    legend cell align={left},
                    legend style={ anchor=north,legend columns=-1},
                    legend to name={legend_ratio_2},
                    xlabel= \#params ($B$),
                    ylabel=\textit{Acc.},
                    ylabel style = {at={(0.06,1.1)},},
                    xticklabel,  
                    enlargelimits = false,
                    xticklabels from table={./model_size_ld.tex}{depth},
                    every axis plot/.append style={ thick},
                    xtick=data]

                    \addplot [ opacity=0.75,mark=triangle*,
                   nodes near coords, color=blue,  error bars/.cd, y dir=both, y explicit,] table [ x=idx, y=LogicCoT, col sep=space, 
                        ]   {./model_size_ld.tex};
                    \addlegendentry{Symbolic-Aided CoT} 
                    \addplot [ opacity=0.75,mark=diamond*, color=red,  error bars/.cd, y dir=both, y explicit,] table [ x=idx, y=CoT, col sep=space, 
                        ]   {./model_size_ld.tex};
                    \addlegendentry{CoT} 
                    \addplot [ opacity=0.75,mark=square*, color=orange,  error bars/.cd, y dir=both, y explicit,] table [ x=idx, y=Standard, col sep=space, 
                        ]   {./model_size_ld.tex};
                    \addlegendentry{Standard} 
                    
                \end{axis} 
            \end{tikzpicture}
            \ref{legend_ratio_2}
            \caption{Performance across different model sizes of \texttt{Qwen-3} with three prompting techniques on the LogicalDeduction dataset. 
            \label{fig_modelsize_ld}}
        \end{figure}

       \paragraph{Impact of model size on performance.} To assess the effectiveness of our Symbolic-Aided CoT across different model sizes, we conducted experiments using various Qwen LLMs on ProofWriter (Figure~\ref{fig_modelsize}) and LogicalDeduction (Figure~\ref{fig_modelsize_ld}) datasets. These results demonstrate that our method consistently outperforms both CoT and standard prompting across model sizes. Furthermore, our approach appears to encourage LLMs to more explicitly articulate the underlying logical reasoning patterns, even in small-scale models. For example, on the ProofWriter dataset, \texttt{Qwen3-8B} achieves performance comparable to that of the \texttt{32B} model. On the LogicalDeduction dataset, \texttt{Qwen3-8B} attains 86.9\% of the performance of the \texttt{32B} model. 
        We argue that our Symbolic-Aided CoT decomposes the original complex logical reasoning tasks into sub-reasoning operations--such as selecting rules, generating new premises, and extending KB premises--that can be effectively addressed by smaller language models.

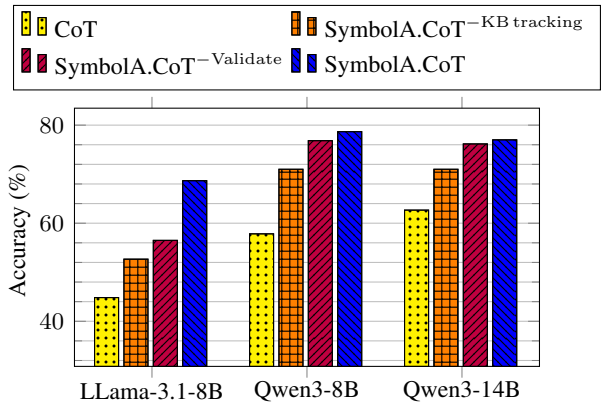
\begin{figure}[htbp]
                \centering
                \begin{tikzpicture}
                \tikzstyle{every node}=[font=\small]
                    \begin{axis} [ybar,
                        width=1.\linewidth,height=0.65\linewidth,
                        bar width = 9pt,
                        ymin=30.8,
                        tick label style={font=\small},
                        ylabel=Accuracy (\%),
                         x label style={at={(current axis.right of origin)},anchor=south, right=4mm, below=1mm}, 
                         y label style={at={(current axis.left of origin)},anchor=south, left=-18mm, below=-10mm}, 
                        ybar=-22, %
                        enlarge x limits = 0.25,
                        symbolic x coords = {LLama-3.1-8B,Qwen3-8B,Qwen3-14B},
                        legend style={
                            at={(0.5,1.4)},
                           anchor=north,
                           legend columns=2,
                           /tikz/every node/.style={anchor=west}
                       },
                        ybar,
                        x tick label style={
                            /pgf/number format/1000 sep=},
                         xticklabel style={rotate=0,anchor=center, yshift=-5pt},
                          minor y tick num=4,
                        ymajorgrids,yminorgrids,
                        ]
                        \addlegendentry{CoT} 
                        \addplot[ybar, bar shift=-17pt, fill=yellow,  postaction={pattern=dots}] coordinates{(LLama-3.1-8B, 44.83) (Qwen3-8B, 57.83) (Qwen3-14B, 62.67)};
                        
                        \addlegendentry{SymbolA.CoT$^{\mathrm{- KB\, tracking}}$} 
                        \addplot[ybar, bar shift=-6pt,  fill=orange,  postaction={pattern=grid}] coordinates{(LLama-3.1-8B, 52.67) (Qwen3-8B,71.00) (Qwen3-14B,71.00)};
                        
                        \addlegendentry{SymbolA.CoT$^{\mathrm{- Validate}}$} 
                        \addplot[ybar, bar shift=5pt,   fill=purple,  postaction={pattern=north east lines}] coordinates{(LLama-3.1-8B, 56.50) (Qwen3-8B,76.83) (Qwen3-14B,76.17)};
                        
                        \addlegendentry{SymbolA.CoT} 
                        \addplot[ybar, bar shift=16pt,  fill=blue,    postaction={pattern=north west lines}] coordinates{(LLama-3.1-8B, 68.67) (Qwen3-8B,78.67) (Qwen3-14B, 77.00)}; 
                    \end{axis} 
                    \end{tikzpicture}
            \caption{Ablation study on the Symbolic-Aided  CoT (SymbolA.CoT).}
            \label{fig_devresult}
            \end{figure}

    \paragraph{Ablation studies.}     
        For evaluating the contribution of each sub-component in our Symbolic-Aided CoT prompting, we conduct two ablation studies: (1) removing the KB-tracking variables (SymbolA.CoT$^{\mathrm{-KB\,tracking}}$), which removes the text segment \textit{``\#KB = {[[KB values for each reasoning step]]}''} in Table~\ref{tab_prompting}, and (2) removing the symbolic $\mathrm{Validate}$ function (SymbolA.CoT$^{\mathrm{-Validate}}$), which removes the text segment \textit{``Validate(Question=[[Q]], KB([[selected premise]]))''} in Table~\ref{tab_prompting}. The ablation results (shown in Figure~\ref{fig_devresult}) indicate that KB-tracking variables play an important role in the reasoning process, helping LLMs avoid loops in the conferencing process. 
        Furthermore, KB-tracking intuitively provides additional features to the hidden representation of premises, allowing the model to distinguish inferred premises from conditional premises in the constraint rules. In another aspect, the symbolic $\mathrm{Validate}$ function in our Symbolic-Aided CoT helps LLMs refer back to the original question in the context to select the appropriate premise for logical matching and producing the final answer.

    \subsection{Result Analysis }
\begin{figure}[!htbp]
            \centering 
            \includegraphics[width=\linewidth, keepaspectratio, 
                    trim={1cm 0.1cm 0.2cm  0}, page=1, clip=true]{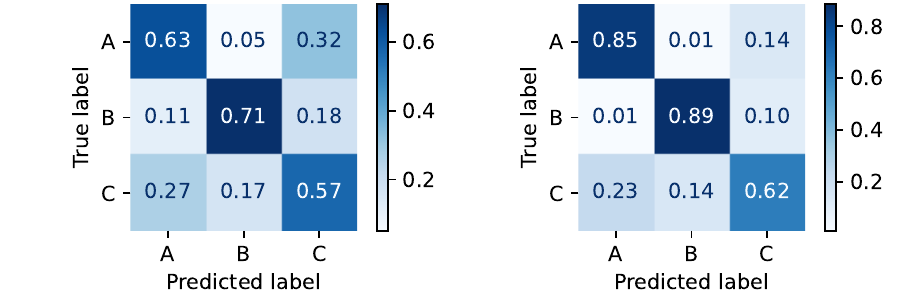}
            \caption{Result comparison with confusion matrices between our Symbolic-Aided CoT (right) and the original CoT (left). The labels A, B, and C refer to the answers True, False, and Uncertain, respectively.  }\label{fig_err_confusionmt}
        \end{figure}

        \paragraph{Confusing ratio.} Here we report the confusion matrices (Figure~\ref{fig_err_confusionmt}) of the answers in the ProofWriter dataset, generated by the \texttt{Qwen-8B} model. For both methods, the original CoT and our Symbolic-Aided CoT, the recall score for False questions is the highest, followed by True and Uncertain. This is due to the complex nature of the logical reasoning task, which involves multi-hop reasoning steps; reasoning paths leading to wrong conclusions are typically more numerous than those leading to correct ones. Comparing our Symbolic-Aided CoT to the original CoT, our method shows improvement across all three question types. The main improvement comes from reducing confusion in Uncertain questions, decreasing misclassification as True or False. We argue that, through symbolic injection, our method encourages clearer logical patterns and structure, thereby enhancing the logical reasoning ability of LLMs.

\begin{figure}[!htbp]
                \centering 
                \includegraphics[width=\linewidth, keepaspectratio, 
                        trim={3cm 1cm 3cm 1cm}, page=1, clip=true]{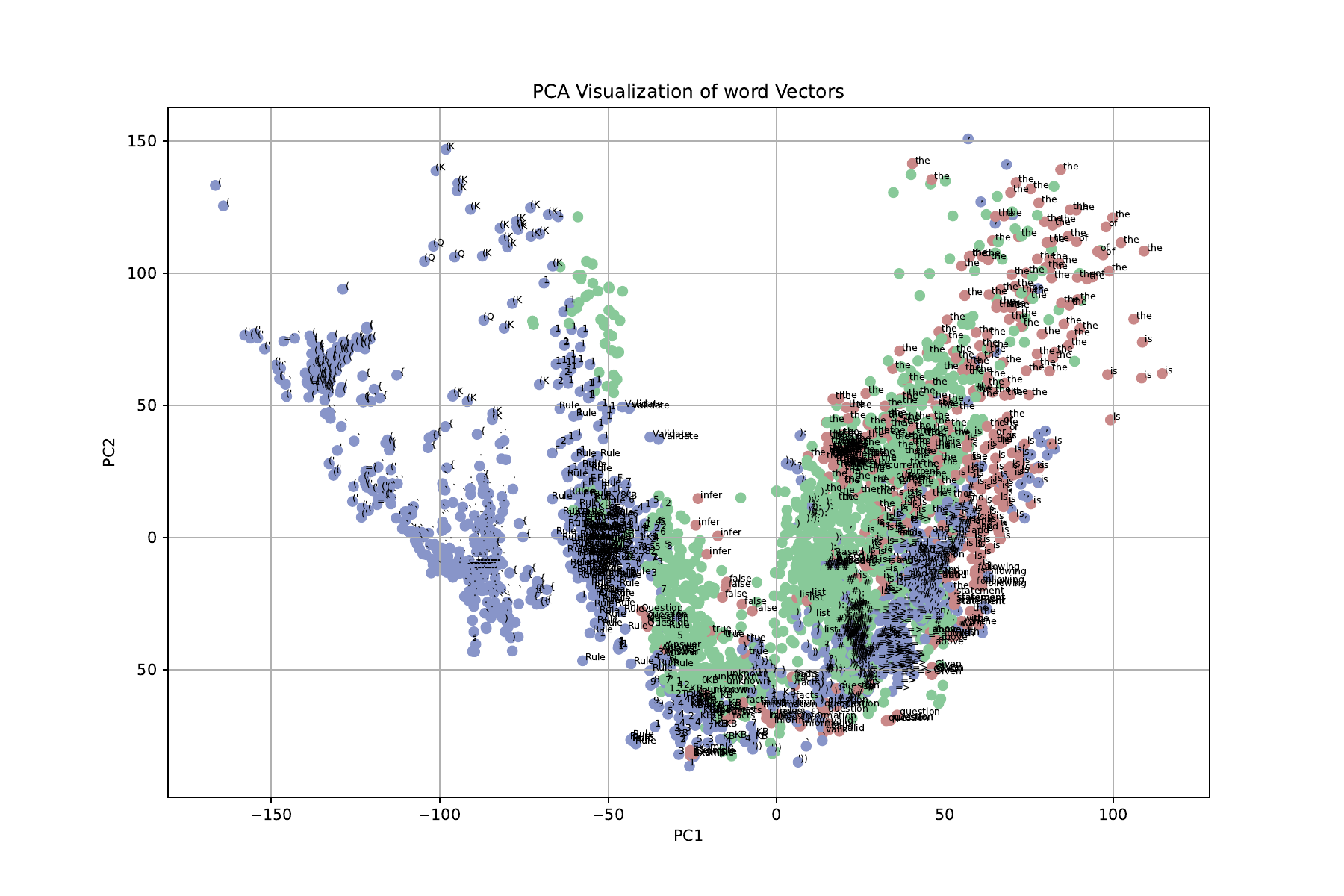}
               
                \caption{Visualization of last-layer word hidden states from \texttt{Qwen3-8B}, with dimensionality reduced via PCA, in the Symbolic-Aided CoT setting on the ProofWriter dataset. The purple, brown, and green data points represent the embeddings of logical symbols (\textit{e.g.,} ``=>'', ``KB''), instruction tokens (\textit{e.g.,} ``Let,'' ``us,'' ``define''), and sample content tokens (\textit{e.g.,} ``cat,'' ''mouse''), respectively.}\label{fig_pca_symbolic}
            \end{figure}

        \paragraph{Semantic representation of symbolic tokens.}
            Here, Figure~\ref{fig_pca_symbolic} visualizes the semantic representation of symbolic tokens using \tien{principal component analysis} (PCA) based on the last layer’s output hidden states of the LLM \texttt{Qwen-3-14B}. This experiment aims to analyze, at a low level, how LLMs understand symbolic tokens in our Symbolic-Aided CoT prompting. We found that LLMs can clearly distinguish the meaning of symbolic tokens (purple data points) from sample content tokens in our proposed prompting method. This is because these logical tokens play the role of structuring the inference flow (latent reasoning pattern) of LLMs, which is separate from the content words in facts and rules. Through few-shot in-context learning, these tokens are represented in a distinct semantic space. Via the self-attention mechanism, logical tokens are paired with content tokens to yield features specific to reasoning operators (such as matching rules or inferring new premises). This suggests that LLMs can uncover the hidden patterns of logical reasoning operators implied by the symbolic tokens within few-shot learning.
               
    
        \paragraph{Case studies.}   Based on case studies of incorrect predictions, we identified several improvement scenarios exhibited by Symbolic-Aided CoT compared to the standard CoT prompting technique (Table~\ref{tab_improving_example} in Appendix \ref{sec:appendix}):
        (1) \textit{Hallucinated Inference Rules:} 
        LLMs often generate inference rules that are either fabricated, logically invalid, or misaligned with the intended knowledge base (first row in Table~\ref{tab_improving_example}). This phenomenon is caused by the counterfactual or pre-trained knowledge embedded within LLMs can conflict with in-context rules or premises provided at inference time. This undermines the assumption that the model will reason strictly within the given context or constraints; (2)	\textit{Unstoppable Inference Flow:}
        The reasoning process lacks a clear halting condition  (third row in Table~\ref{tab_improving_example}). The model continues generating premises without a mechanism to determine when inference should stop, leading to uncontrolled or incoherent inference chains. This highlights the need to explicitly track and manage the state of the knowledge base (KB) during reasoning; (3) \textit{Failure on Cyclic Inference Graphs:} When the inference space forms a cyclic graph, LLMs often fail—either entering infinite reasoning loops or struggling to resolve the cycle  (second and third rows in Table~\ref{tab_improving_example}). These models lack the structural awareness to detect and handle loops in reasoning chains; (4) \textit{Rule Matching Errors:} LLMs frequently fail to apply inference rules correctly due to poor condition matching  (fourth row in Table~\ref{tab_improving_example}). The model may skip necessary preconditions or generate incorrect intermediate steps, breaking the logical flow of multi-step reasoning.
        
\section{Discussion\label{sec_discusion}}        
Logical reasoning tasks have attracted numerous research works recently~\citep{pan-etal-2023-logic, ye2023satlm, olausson2023linc, sun-etal-2024-determlr, zhang2025cumulative}, especially following the massive success of LLMs. Unlike previous methods, by proposing Symbolic-Aided CoT, we primarily aim to enhance the logical reasoning ability of LLMs rather than simply build a system to improve performance on logical reasoning tasks. For example, frameworks such as Logic-LM~\cite{pan-etal-2023-logic} and SatLM~\cite{ye2023satlm} use LLMs only to translate logical problems into inputs for explicit symbolic reasoners. Frameworks like CR~\cite{zhang2025cumulative} and DetermLR~\cite{sun-etal-2024-determlr} leverage LLMs to perform small constituent logical reasoning steps, rather than directly evaluating the LLM’s logical reasoning ability on the entire problem. 

The experimental results show that our Symbolic-Aided CoT prompting technique is reliable and effectively improves the logical reasoning ability of LLMs, even for small model sizes. Our prompting method is simple, yet effective and flexible, allowing customization for any logical reasoning task. It can also yield proof trees that facilitate explanation and enhance transferability.
           
\section{Conclusion\label{sec_conclusion}}
In this work, we introduced Symbolic-Aided CoT, a novel prompting technique for \phuong{non-interactive} logical reasoning, which achieves superior performance on well-known benchmark datasets—most notably ProofWriter, ProntoQA, and LogicalDeduction. Our method is deliberately simple to preserve generalizability and shows strong potential for extension to other reasoning tasks. For future work, Symbolic-Aided CoT, grounded in structural characteristics, could be combined with mechanisms for refining the latent semantic vector space, thereby further improving the faithfulness and reliability of LLMs’ reasoning capabilities.

\section*{Acknowledgments}

This work was supported by JSPS KAKENHI Grant Number 22H00524 and the Nakajima Foundation. We used ABCI 3.0 provided by AIST and AIST Solutions with support from ``ABCI 3.0 Development Acceleration Use''.
 
\section*{Limitations}
We discuss the following limitations and future works: We have evaluated the proposed method on four widely used logical reasoning benchmarks. However, they are mostly synthetic; incorporating real-world datasets or diverse reasoning tasks (\textit{e.g.,} commonsense reasoning) would strengthen claims of generalizability. Relying solely on automatic metrics like accuracy overlooks qualitative aspects; integrating human evaluations to assess reasoning faithfulness and interpretability would offer a more holistic validation. Future research could also explore the method’s robustness to adversarial perturbations, sensitivity to prompts, and scalability to longer reasoning chains.
\bibliography{ref}

\appendix
\section{Improving examples\label{sec:appendix}}
In this section, we present detailed improvement examples from the ProofWriter dataset for our Symbolic-Aided CoT method compared to the CoT method, in the setting using the \texttt{Qwen3-8B} model (Table~\ref{tab_improving_example}).

         \begin{table*}[!htbp]
                \small
                    \caption{{Improving examples from the ProofWriter dataset for our Symbolic-Aided CoT method compared to the CoT method. The red text refers to incorrect logical reasoning or hallucinated content, and the green text refers to the final premises related to the target question.}\label{tab_improving_example}}
                \centering
                \resizebox{\textwidth}{!}{%
                    \begin{tabular}{p{0.1\linewidth}p{0.45\linewidth}p{0.35\linewidth}p{0.4\linewidth}ll}
                        %
                        \toprule  \multirow{1}{*}{\textbf{Id}}&\textbf{Context, Question, and Answer}&\textbf{CoT's output } + \noindent\colorbox{gray1}{\fbox{ %
                        \textbf{\textit{Error Type}   }}} &\textbf{Symbolic-Aided CoT's output} + \noindent\colorbox{gray1}{\fbox{ %
                        \textbf{\textit{Error Type}   }}}  \\\midrule
ProofWriter-\_AttNeg-OWA-D5-596\_Q6
&  
\#\#\# Given list of facts and rules: \newline
\# (Rule1): Anne is not rough. \newline
\# (Rule2): Bob is blue. \newline
\# (Rule3): Erin is not furry. \newline
\# (Rule4): Erin is red. \newline
\# (Rule5): Gary is big. \newline
\# (Rule6): Gary is not cold. \newline
\# (Rule7): Gary is furry. \newline
\# (Rule8): If something is blue then it is rough. \newline
\# (Rule9): Red things are rough. \newline
\# (Rule10): If Gary is rough then Gary is not blue. \newline
\# (Rule11): Rough things are red. \newline
\# (Rule12): Big things are quiet. \newline
\# (Rule13): All cold things are big. \newline
\# (Rule14): If something is red then it is big. [...] \newline
\# (Question): Based on the above information, is the following statement true, false, or unknown? Erin is not quiet. \newline
\# (Answer): \textbf{False}
& 
 Erin is red. Red things are rough. \textcolor{red}{If something is rough then it is not quiet}. So Erin is not quiet.  \,\, \, \, \, \, \, \, \, \, \, \, \, \, \, \, \, \,  
\noindent\colorbox{gray1}{\fbox{\parbox{0.95\linewidth}{\textit{%
                      (1) Hallucinated Inference Rules 
}}}}
& 
Start from the object and their condition mentioned in the question to collect relevant facts: Erin, is not quiet\newline
\# KB = \{\} \newline
=> Rule3 = `Erin is not furry`\newline
=> Rule4 = `Erin is red`\newline
\# KB = \{Erin is not furry, Erin is red\} \newline
=> F(KB('Erin is red'), Rule9) => `Erin is rough`\newline
\# KB = \{Erin is not furry, Erin is red, Erin is rough\} \newline
=> F(KB('Erin is red'), Rule11) => `Erin is red` (already in KB)\newline
\# KB = \{Erin is not furry, Erin is red, Erin is rough\} \newline
=> F(KB('Erin is red'), Rule14) => `Erin is big`\newline
\# KB = \{Erin is not furry, Erin is red, Erin is rough, Erin is big\} \newline
\textcolor{ForestGreen}{=> F(KB('Erin is big'), Rule12) => `Erin is quiet`}\newline
\# KB = \{Erin is not furry, Erin is red, Erin is rough, Erin is big, \textcolor{ForestGreen}{Erin is quiet}\} \newline
\# valid the question with current inferred premies\newline
=> Validate(Question=`Erin is not quiet`, KB('Erin is quiet'))=\textbf{False}.\\  \midrule

                       ProofWriter-\_RelNoneg-OWA-D5-356\_Q4 
& 
\#\#\# Given list of facts and rules:\newline
\# (Rule1): The cow eats the mouse.[...]\newline
\# (Rule15): The squirrel likes the cow.\newline
\# (Rule16): The squirrel likes the mouse.\newline
\# (Rule17): If something likes the cow then it visits the mouse.\newline
\# (Rule18): If something likes the cow then the cow is green.\newline
\# (Rule19): If something eats the squirrel and the squirrel likes the rabbit then it eats the rabbit.\newline
\# (Rule20): If something is cold then it likes the cow.\newline
\# (Rule21): If something visits the squirrel then the squirrel likes the rabbit.\newline
\# (Rule22): If the cow likes the rabbit then the cow is cold.[...]\newline
\# (Question): Based on the above information, is the following statement true, false, or unknown? The squirrel does not visit the mouse.\newline
\# (Answer): \textbf{False}& The cow likes the rabbit. If the cow likes the rabbit then the cow is cold. The cow is cold. The cow likes the squirrel. If something is cold then it likes the cow. The cow likes the cow. \textcolor{red}{The cow visits the mouse}. If something likes the cow then it visits the mouse. \textcolor{red}{The cow visits the mouse}. The squirrel likes the cow. If something visits the squirrel then the squirrel likes the rabbit. The squirrel likes the rabbit. \textcolor{red}{The squirrel eats the rabbit.} If something likes the rabbit and it visits the mouse then the mouse visits the squirrel [...] \newline 
\noindent\colorbox{gray1}{\fbox{\parbox{0.95\linewidth}{\textit{%
(3) Failure on Cyclic Inference Graph \newline
(4) Rule Matching Errors
}}}}
& 
Start from the object and their condition mentioned in the question to collect relevant facts: The squirrel, does not visit the mouse\newline
    \# KB = \{\} \newline
    => Rule1 = `The cow eats the mouse`\newline
    \# KB = \{cow eats mouse, [...], squirrel is rough, squirrel likes cow, squirrel likes mouse\} \newline
    \textcolor{ForestGreen}{=> F(KB('squirrel likes cow'), Rule17) => `squirrel visits mouse`}\newline
    \# KB = \{cow eats mouse, [...] squirrel is rough, squirrel likes cow, squirrel likes mouse, \textcolor{ForestGreen}{squirrel visits mouse}\} \newline
    \# valid the question with current inferred premies\newline
    => Validate(Question=`The squirrel does not visit the mouse`, KB('squirrel visits mouse')) = \textbf{False}.\\ \midrule
    
                       ProofWriter-\_RelNeg-OWA-D5-688\_Q22
& 
\#\#\# Given list of facts and rules:[...]\newline
\# (Rule6): The lion needs the mouse.\newline
\# (Rule7): The mouse is round.\newline
\# (Rule8): The mouse likes the lion.\newline
\# (Rule9): The mouse needs the lion.[...]\newline
\# (Rule15): If something likes the cat and it does not visit the cat then it visits the lion.\newline
\# (Rule16): If the lion is green and the lion visits the mouse then the mouse is red.\newline
\# (Rule17): All red things are nice.\newline
\# (Rule18): If something likes the lion then it is red.\newline
\# (Rule19): If the mouse visits the cat and the mouse needs the squirrel then the mouse does not like the cat.\newline
\# (Rule20): Nice things are big.\newline
\# (Rule21): If something is big then it visits the squirrel.\newline
\# (Rule22): If the mouse visits the squirrel then the mouse does not visit the lion.\newline
\# (Question): Based on the above information, is the following statement true, false, or unknown? The lion likes the mouse.\newline
\# (Answer):": \textbf{Uncertain}
& 
The mouse likes the lion. The mouse needs the lion. The mouse visits the cat. The mouse visits the squirrel. The mouse does not visit the lion. The mouse is round. The mouse is red. The mouse is nice. The mouse is big. The mouse is nice. The mouse is big. The mouse visits the squirrel. The mouse does not visit the lion. \textcolor{red}{The mouse is red. The mouse is nice. The mouse is big. The mouse is nice. The mouse is big. The mouse is nice. The mouse is big. The mouse is nice. The mouse is big.  } [...] \newline 
\noindent\colorbox{gray1}{\fbox{\parbox{0.95\linewidth}{\textit{%
                      (2) Unstoppable Inference Flow \newline
                      (3) Failure on Cyclic Inference Graphs
}}}}
& 
Start from the object and their condition mentioned in the question to collect relevant facts: The lion, likes the mouse\newline
\# KB = \{\} [...]\newline
\# KB = \{The lion needs the mouse, The mouse likes the lion, The mouse is red, The mouse is nice\} \newline
=> F(KB('The mouse is nice'), Rule20) => `The mouse is big`\newline
\# KB = \{The lion needs the mouse, The mouse likes the lion, The mouse is red, The mouse is nice, The mouse is big\} \newline
=> F(KB('The mouse is big'), Rule21) => `The mouse visits the squirrel`\newline
\# KB = \{The lion needs the mouse, The mouse likes the lion, The mouse is red, The mouse is nice, The mouse is big, The mouse visits the squirrel\} \newline
=> F(KB('The mouse visits the squirrel'), Rule22) => `The mouse does not visit the lion`[...]\newline
\# valid the question with current inferred premies\newline
=> Validate(Question=`The lion likes the mouse`, KB('The mouse likes the lion')) =\textbf{Uncertain}.\\  \midrule
    
                       ProofWriter-\_AttNoneg-OWA-D5-245\_Q19
& 
\#\#\#  Given list of facts and rules:[...]\newline
\# (Rule4): Erin is nice.[...]\newline
\# (Rule11): Furry people are smart.\newline
\# (Rule12): If someone is smart and nice then they are round.\newline
\# (Rule13): Cold people are red.\newline
\# (Rule14): If someone is quiet then they are nice.\newline
\# (Rule15): All red people are furry.\newline
\# (Rule16): All smart, furry people are nice.\newline
\# (Question): Based on the above information, is the following statement true, false, or unknown? Erin is not quiet. \newline
\# (Answer) : \textbf{Uncertain}
& 
Erin is nice. If someone is nice then they are quiet. \textcolor{red}{So Erin is quiet.} Therefore, Erin is quiet, so the statement 'Erin is not quiet' is false.   \newline 
\noindent\colorbox{gray1}{\fbox{\parbox{0.95\linewidth}{\textit{%
                      (4) Rule Matching Errors
}}}}
&
Start from the object and their condition mentioned in the question to collect relevant facts: Erin, is not quiet\newline
\# KB = \{\} \newline
=> Rule4 = `Erin is nice`\newline
\# KB = \{Erin is nice\} \newline
\textcolor{red}{=> F(KB('Erin is nice'), Rule14) => `Erin is quiet`}\newline
\# KB = \{Erin is nice, Erin is quiet\} \newline
\# valid the question with current infered premies\newline
=> Validate(Question=`Erin is not quiet`, KB('Erin is quiet')) = \textbf{False}.\newline
\noindent\colorbox{gray1}{\fbox{\parbox{0.95\linewidth}{\textit{%
                      (4) Rule Matching Errors
}}}}
\\  
    
                        \bottomrule 
                    \end{tabular} 
                    } 
        \end{table*}
        

\end{document}